%% file: acl_latex.tex
\pdfoutput=1

\documentclass[11pt]{article}

\usepackage{acl}

\usepackage{times}
\usepackage{latexsym}

\usepackage[T1]{fontenc}

\usepackage[utf8]{inputenc}

\usepackage{microtype}
\usepackage{graphicx}
\usepackage{amsmath,amssymb,bm}
\usepackage{xspace}
\usepackage{tabularx}
\usepackage{booktabs,arydshln}
\usepackage{tabu}

\usepackage{subcaption}
\usepackage{colortbl}
\usepackage{makecell}
\usepackage{float}
\usepackage{enumitem}
\usepackage[ruled,vlined]{algorithm2e}
\usepackage{smile}

\newcommand{\ours}{{{\textsc{SeqZero}}}} 

\title{{\ours}: Few-shot Compositional Semantic Parsing with \\ Sequential Prompts and Zero-shot Models}

\author{First Author \\
  Affiliation / Address line 1 \\
  Affiliation / Address line 2 \\
  Affiliation / Address line 3 \\
  \texttt{email@domain} \\\And
  Second Author \\
  Affiliation / Address line 1 \\
  Affiliation / Address line 2 \\
  Affiliation / Address line 3 \\
  \texttt{email@domain} \\}
  
\author{Jingfeng Yang$^{\dagger}$ \quad Haoming Jiang$^{\dagger}$ \quad Qingyu Yin$^{\dagger}$\\
\bf{Danqing Zhang}$^{\dagger}$ \quad Bing Yin$^{\dagger}$ \quad Diyi Yang$^{\ddagger}$\\
  $^{\dagger}$ Amazon\\
  $^{\ddagger}$ Georgia Institute of Technology\\
  {\tt \{jingfe, jhaoming, qingyy, danqinz, alexbyin\}@amazon.com} \\  {\tt dyang888@gatech.edu } \\
  }

\begin{document}
\maketitle

\input{0_abstract}
\input{1_intro}
\input{3_method}

\input{4_exp}
\input{2_relatedwork}
\input{5_conclusion}

\bibliography{anthology,custom}
\bibliographystyle{acl_natbib}

\appendix

\input{appendix}

\end{document}

%% file: 0_abstract.tex
\begin{abstract}
Recent research showed promising results on combining pretrained language models (LMs) with canonical utterance for few-shot semantic parsing.
The canonical utterance is often lengthy and complex due to the compositional structure of formal languages. 
Learning to generate such canonical utterance requires significant amount of data to reach high performance. Fine-tuning with only few-shot samples, the LMs can easily forget pretrained knowledge, overfit spurious biases, and suffer from compositionally out-of-distribution generalization errors. 
To tackle these issues, we propose a novel few-shot semantic parsing method -- {\ours}. {\ours} decomposes the problem into a sequence of sub-problems, which correspond to the sub-clauses of the formal language. 
Based on the decomposition, the LMs only need to generate short answers using prompts for predicting sub-clauses. Thus, {\ours} avoids generating a long canonical utterance at once. 
Moreover, {\ours} employs not only a few-shot model but also a zero-shot model to alleviate the overfitting.
In particular, {\ours} brings out the merits from both models via ensemble equipped with our proposed constrained rescaling.
{\ours} achieves SOTA performance of BART-based models on GeoQuery and  EcommerceQuery, which are two few-shot datasets with compositional data split.\footnote{Code and data to be released at \url{https://github.com/amzn/SeqZero.}} 

\end{abstract}



%% file: 1_intro.tex
\section{Introduction}
\label{sec:intro}

Semantic parsing is the transformation of input utterance into formal language, such as SQL query \citep{zelle1996learning},  and plays a critical role in NLP applications, such as question answering \citep{yih2014semantic}, dialogue system \citep{gupta2018semantic}, and information extraction \citep{yao2014information}. 
Training neural semantic parsers requires numerous annotated input utterance and formal language pairs. 
However, the paired data is usually limited, as the annotation requires experts’ knowledge and can be expensive. For example, annotating SQL queries requires programming knowledge, while annotating formal meaning representations like Abstract Meaning Representations (AMR) requires linguistics knowledge. Therefore, semantic parsing in the few-shot setting is a demanding technique.

Researchers have adopted large-scale pretrained language models (LMs, \citet{radford2019language,brown2020language}) to improve few-shot learning performance. The LMs are usually pretrained on large unlabeled open-domain natural language data and achieve impressive performance on few-shot text-to-text generation problems via proper prompt designing \citep{brown2020language}. Considering the difference between natural and formal language, adapting LMs to semantic parsing is non-trivial. Prior works typically first finetune the LMs to generate canonical utterance, which is then transformed into the final formal language through grammars \cite{shin2021constrained, schucher2021power}.

\begin{figure}
\begin{center}
\includegraphics[width=0.7\linewidth]{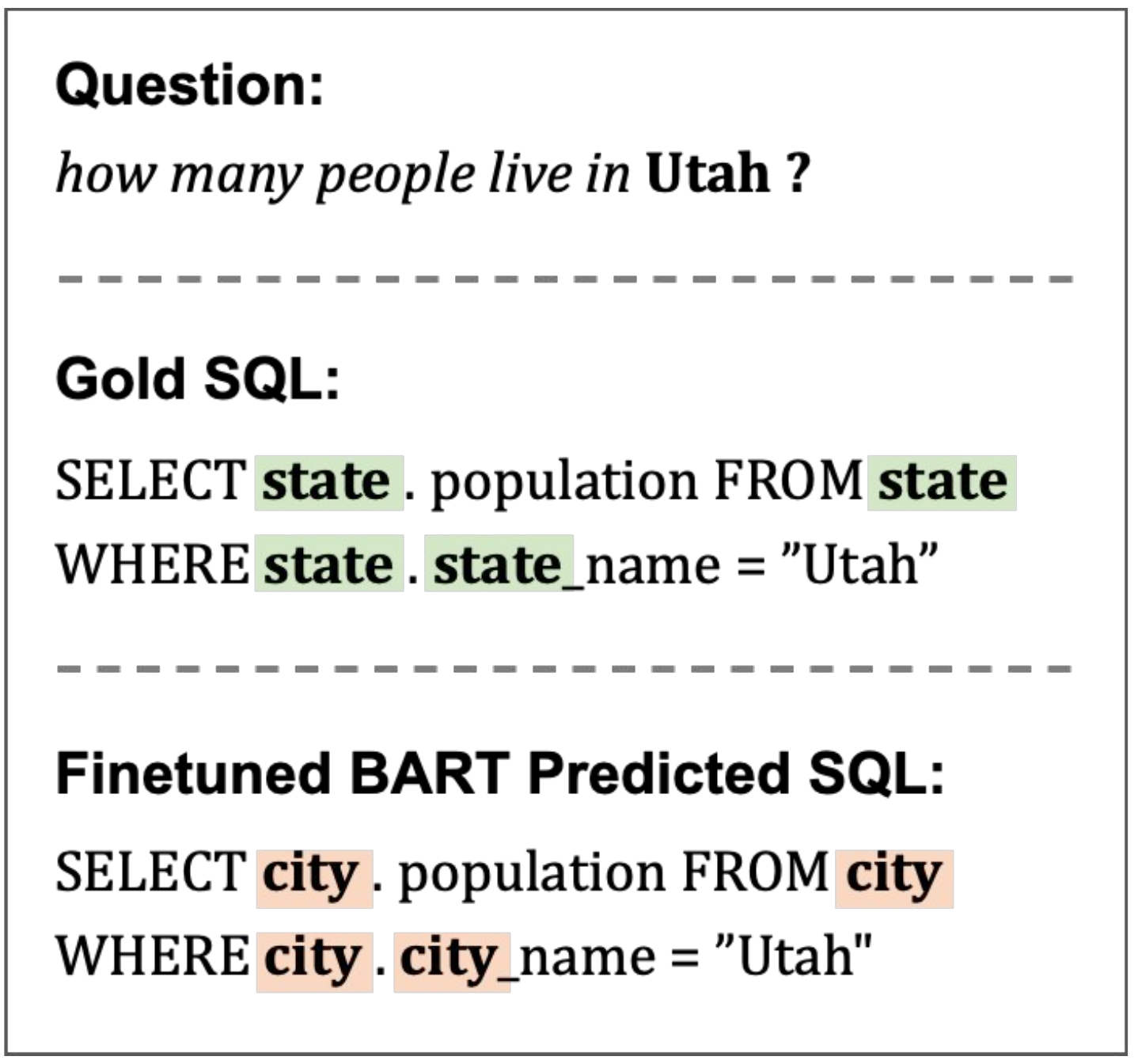}
\end{center}
\caption{\label{fig:example}  Finetuned BART's OOD generalization errors due to overfitting the spurious biases.
}
\end{figure}

However, the canonical utterance is lengthy and complex due to compositional structure of the formal languages. Learning to precisely generate canonical utterances still requires significant amount of data. Meanwhile, fine-tuning with only few-shot samples, the LMs can easily forget pretrained knowledge, overfit spurious biases, and suffer from compositionally out-of-distribution (OOD) generalization errors. 
Figure \ref{fig:example} presents an compositionally OOD generalization error of direct fine-tuning BART \citep{lewis2019bart} on the GeoQuery, a dataset about querying in a geographic database. The model incorrectly predicts the table name as ``\textit{city}'', because the training samples always come from the ``\textit{city}'' table as long as the query follows the ``\textit{how many people live in xxx}'' pattern. Such errors account for about 75\% of all prediction errors of Base model on GeorQuery test set (refer to Section~\ref{case-study} for details). 


To address the aforementioned issues, we propose a novel prompt-based few-shot learning method -- {\ours}. 
Instead of directly generating the whole formal language, {\ours} decomposes the problem into a sequence of sub-problems, and the LMs only need to make a sequence of short prompt-based predictions, where zero-shot (un-finetuned) models can also be leveraged to avoid overfitting the spurious biases in specific caluses.
Specifically, {\ours} decomposes the problem into predicting the sub-clauses, which make up the formal languages. When predicting a sub-clause, {\ours} adopts a slot-filling natural language prompt, where the filled prompt can be transformed into the sub-clause through grammars. 
For filling each prompt, {\ours} employs two models: a few-shot model and a zero-shot model. Both models ingest the input utterance and the prompt to fill in the slots in the prompt. 
The few-shot model uses a fine-tuned LM to fill in the slots of each prompt. The zero-shot model directly infers the value in the slots by decoding a pretrained LM with a constrained vocabulary. We then ensemble the prediction from both models, and convert the results for all sub-clauses into the final output (e.g., SQL query). 
We notice that, the probability mass of the zero-shot model, on the constrained vocabulary, is much smaller than that of the few-shot model. As a result, the zero-shot model cannot take effect in the vanilla ensemble.
Therefore, we propose to rescale the probability of the zero-shot model on the constrained vocabulary before ensemble to bring out the advantages of both models.  

We conduct experiments on two datasets: GeoQuery, a benchmark dataset that consists of natural language and formal language pairs from geography domain, and EcommerceQuery, a newly collected dataset from E-commerce domain. Results show that our approach outperforms the baseline algorithm and achieves state-of-the-art performance on the compositional split of the two datasets.
To sum up, our contributions are: 
\begin{itemize} \setlength\itemsep{0em}
    \item We propose to decompose semantic parsing to filling a sequence of prompts, each corresponding to a sub-clause of original SQL query. 
Compared with direct fine-tuning, predicting sub-clauses is easier, which enables flexible prompt designing and zero-shot model inference. 
\item We propose the ensemble of few-shot and zero-shot models with help of constrained probability rescaling, which improves out-of-distribution generalization while maintaining in-distribution performance. 
\item We create and release a new EcommerceQuery dataset. We empirically verify that our approach achieves SOTA, among BART-based models, on both GeoQuery and EcommerceQuery.

\end{itemize}

%% file: 3_method.tex
\section{Preliminary}
\noindent\textbf{Language Modeling} aims to estimate the probability distribution for a given sequence of words $x=(w_1, w_2, ..., w_n)$ in an autoregressive way:
\begin{align*}
    P_{\theta}(x) = \prod_{i=1}^n P_{\theta}(w_i | w_1, ..., w_{i-1}),
\end{align*}
where $\theta$ is the parameters of the language model. 
This approach not only allows estimation of $P_{\theta}(x)$ but also any conditionals of the form $P_{\theta}(w_i, w_{i+1}, .., w_n | w_1, ..., w_{i-1})$, which is essentially a seq2seq model. One can leverage a seq2seq model to generate a sequence via a decoding algorithm (e.g., beam-search): $y={\rm Decode} (P_\theta(\cdot| x))$
In recent years, there have been significant progress in training large transformer-based language models \citep{radford2019language, brown2020language, lewis2019bart} on large natural language corpus.

\noindent\textbf{Semantic Parsing} is to transform an input utterance $u$ into a formal language $m$. Without loss of generality, we hereafter discuss the case of SQL query as the formal language. One can directly train a language model for semantic parsing:
\begin{align*}
    P_{\theta} (m|u).
\end{align*}

Directly learning such a language model is challenging as the difference between the formal language and natural language is huge. To bridge the gap, \citet{berant2014semantic, shin2021constrained} propose \textbf{S}emantic \textbf{P}arsing via \textbf{P}araphrasing (SPP) --- a two-stage framework. In the first stage, they paraphrase $u$ to its canonical utterance $c$ using a paraphrasing language model:
\begin{align*}
    P_{\theta} (c|u).
\end{align*}

In the second stage, the canonical utterance $c$ is transformed into SQL query $m$ by a grammar or a set of rules:
\begin{align*}
    m = {\rm Grammar}(c).
\end{align*}



\section{Method}

\begin{figure}[!bth]
\begin{center}
\includegraphics[width=0.5\textwidth]{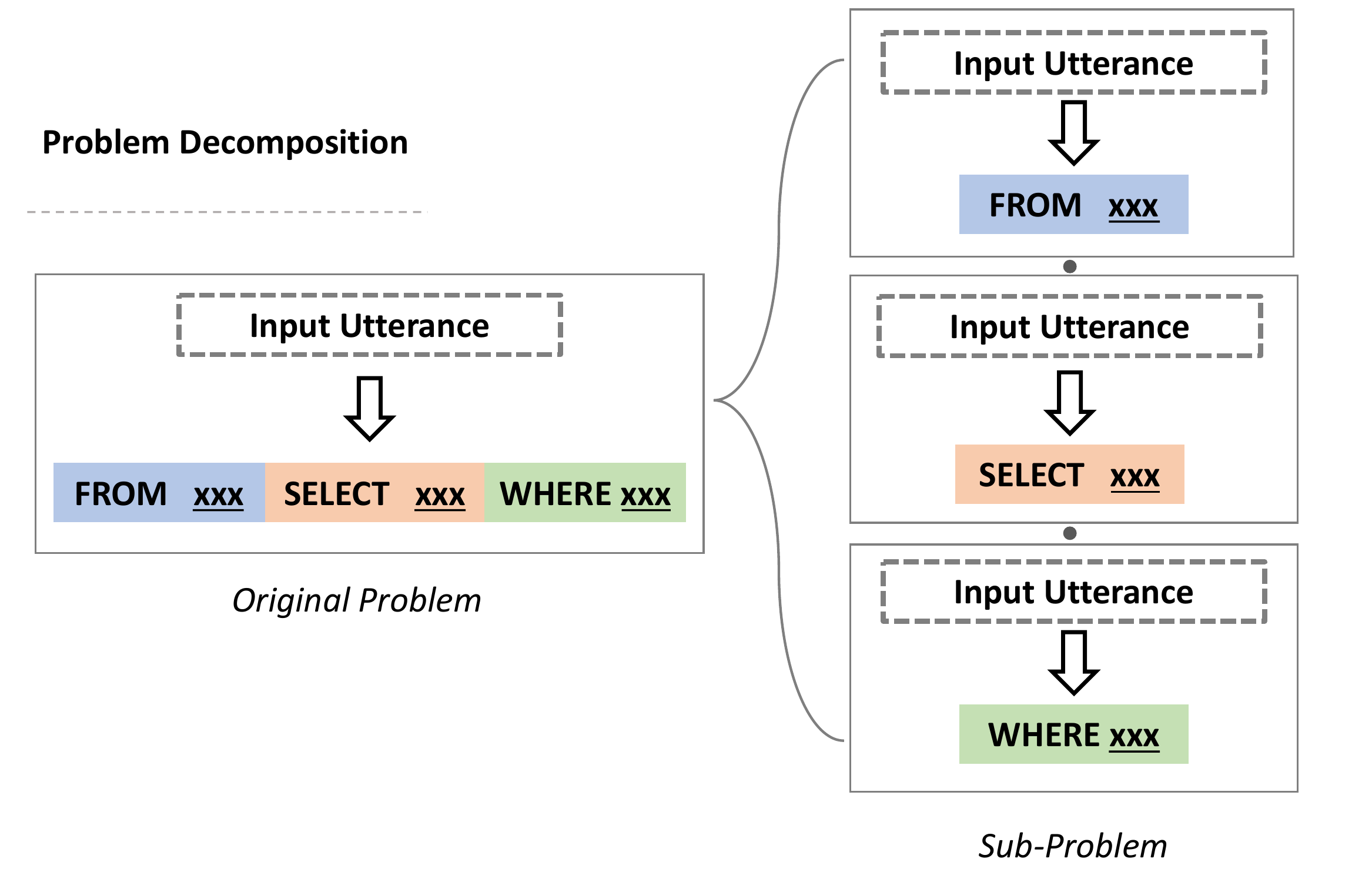} 
\end{center}
\caption{The problem of predicting a SQL can be composed into 3 steps: predicting ``FROM'' clause, ``SELECT'' clause, and ``WHERE'' clause.}
\label{fig:decompose}
\end{figure}

\begin{figure*}
\begin{center}
\includegraphics[width=1\linewidth]{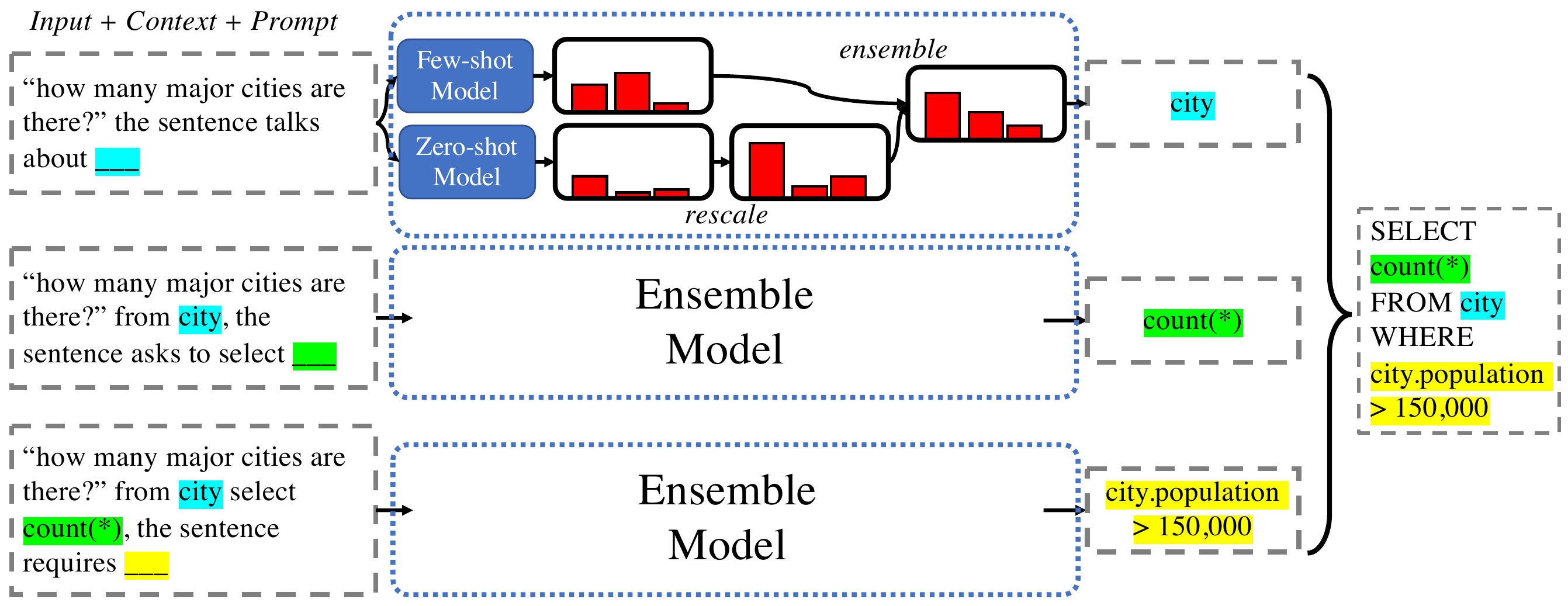} 
\end{center}
\caption{Pipeline of sequential prompt filling and SQL generation on GeoQuery. Note that, the scale of the prediction probability of the zero-shot model is very small before rescaling. 
\label{fig:pipeline}
}
\end{figure*}

In this section, we describe {\ours}. {\ours} first decomposes the problem into a sequence of sub-problems as illustrated in Figure~\ref{fig:decompose}. 
For each sub-problem, {\ours} employs an ensemble of zero-shot and few-shot models to predict a sub-clause of the formal language based on prompts as illustrated in Figure~\ref{fig:pipeline}.

\subsection{Problem Decomposition and Sequential Prompt Filling}

Each SQL query can be regarded as a composition of different types of sub-clauses, such as ``SELECT'', ``FROM'', ``WHERE'':
\begin{align*}
    m = {\rm Compose}(m_1, ..., m_n),
\end{align*}
where $m_i$ is the sub-clause of the $i$-th type, $n$ is the number of all possible types of sub-clauses, and the composition is conducted via a rule-based system. A simple example of the composition function is direct concatenating the sub-clauses, whereas the real implementation requires some dedicated design. For example, $m_i$ can be a null clause, e.g., not every SQL query contains a ``WHERE'' clause. We discuss the implementation details of the composition in Appendix~\ref{app:compose}.

We turn the problem of direct predicting $m$ into predicting $m_i$ sequentially from $m_1$ to $m_n$. We remark that the prediction of $m_i$ depends on $m_1, ..., m_{i-1}$, as illustrated in Figure~\ref{fig:pipeline}. Similar to the SPP framework, we design a canonical utterance $c_i$ for each sub-clause $m_i$. The transformation between $c_i$ and $m_i$ is conducted by a grammar:
\begin{align*}
    m_i = {\rm Grammar} (c_i) .
\end{align*}
Each $c_i$  consists of two parts: a natural language slot-filling prompt $p_i$ and a value in the slot $v_i$: 
\begin{align*}
    c_i = {\rm FillSlot}(p_i, v_i) .
\end{align*}

The prompt $p_i$ is shared across all sub-clauses of the $i$-th type, while the value $v_i$ varies for different instances. As a result, the problem is turned into predicting the values $\{v_i\}_{i=1}^n$ given the input utterance $u$, and prompts $\{p_i\}_{i=1}^n$ sequentially from $i=1$ to $i=n$. The prediction is conducted via decoding a language model, $P_{\theta_i} (\cdot | u, m_1, \dots, m_{i-1}, p_i)$, where the canonical utterances of previous sub-clauses $(m_1, \dots, m_{i-1})$ are also provided as the extra context. We summarize the process in Algorithm~\ref{alg:seq_prompt}.

\begin{algorithm}[htb!]
\SetAlgoLined
\caption{Sequential Prompt Filling}
\label{alg:seq_prompt}
\KwIn{$u$: input utterance;      
$\{p_i\}_{i=1}^n$: prompts;
${\rm Grammar}$: grammar for parsing the canonical utterance; $\{P_{\theta_i}\}_{i=1}^n$: LMs.}
\For{$i = 1, \cdots, n$}{
    $x = (u, m_1, \dots, m_{i-1}, p_i) $ \\
    $v_i = {\rm Decode}( P_{\theta_i} (\cdot | x) )$ \\
    $c_i = {\rm FillSlot}(p_i, v_i)$ \\
    $m_i = {\rm Grammar} (c_i)$ 
}
$m = {\rm Compose}(m_1, ..., m_n)$ \\
\KwOut{$m$: SQL query}
\end{algorithm}

\subsection{Ensemble of Few-shot and Zero-shot Models}
\label{few-shot-model}

Despite the apparent advantages of sequential prompt filling, directly fine-tuning LMs on few-shot samples will fall short due to the overfitting.
Because of the better OOD generalizability of zero-shot models, we propose to employ the ensemble of a few-shot model $P_{\theta_{i,f}}$ and a zero-shot model $P_{\theta_{i,z}}$ for each language model $P_{\theta_{i}}$. 

\noindent\textbf{Few-shot Model}. Each few-shot model is obtained by finetuning a pretrained language model via minimizing the negative log-likelihood loss:

\begin{align*}
\arg\min_{\theta_{i,f}} - \log P_{\theta_{i,f}} (v_i | u, m_1, \dots, m_{i-1}, p_i), 
\end{align*}
where $v_i, m_1, \dots, m_{i-1}$ are the ground truth from the few-shot training data. It is essentially the teacher forcing training strategy. Note that we omit the summation over the training set for simplicity and clarity. 

\noindent\textbf{Zero-shot Model}. Each zero-shot model directly adopts the pretrained language model $P_{\theta_{0}}$. Without any guidance, $P_{\theta_{0}}$ may generate any free text even if we provide the input utterance and prompt. In order to mine the knowledge from $P_{\theta_{0}}$, we only allow the zero-shot model to generate from a list of candidate values. 
The candidate values are collected from multiple sources including SQL grammar, table schema, input utterance and training data.  
When predicting the $j$-th word 
for $v_i$, the zero-shot model rescales the probability on a constraint vocabulary, which is specifically designed for the $i$-th clause:
\begin{align}
    P_{\theta_{i,z}} (w|x) = \frac{\ind(w \in V_i(x)) P_{\theta_{0}} (w|x)}{\sum_{w_j \in V_i(x)} P_{\theta_{0}} (w_j|x) },
    \label{eq:zero-model}
\end{align}
where $w$ is a predicting word, $x=(u,m_1,...,m_{i-1},p_i, w_1, .., w_{j-1})$ is the context for predicting the $i$-th value, $\{w_t\}_{t=1}^{j-1}$ is the prefix in the value, $V_i(x)$ is the constraint vocabulary. Given the list of candidate values, we use a trie (prefix tree) to compute all the allowed tokens, and thus $V_i(x) = V_i(\{w_t\}_{t=1}^{j-1})$ depends on the prefix of the values. 
Note that, to develop a more flexible method, a trie/prompt could start at intermediate steps. 

\noindent\textbf{Ensemble}. 
We then obtain $P_{\theta_{i}}$ by a linear ensemble of the few-shot model $P_{\theta_{i,f}}$ and the zero-shot model $P_{\theta_{i,z}}$:

\begin{align}
    P_{\theta_{i}} = \gamma_i P_{\theta_{i,f}} + (1-\gamma_i)P_{\theta_{i,z}}, 
    \label{eq:ensemble}
\end{align}
where $\gamma_i$ is a clause-specific weight for trade-off between two models.

\textbf{Remark}. We employ a normalization step in the zero-shot model Eq.~\eqref{eq:zero-model}. The normalization is not necessary for the zero-shot model itself, but plays a critical role in the ensemble. 
This is because the scales of the predicted probabilities of few-shot and zero-shot models are different, as illustrated in Figure~\ref{fig:pipeline}.
The $P_{\theta_0}$'s prediction probability is distributed over the whole vocabulary. There is only a very small probability mass assigned to the allowed tokens, $V_i(x)$. On the other hand, the few-shot model's prediction probability is almost entirely distributed over $V_i(x)$. Without rescaling, the zero-shot model will only have little effect when ensembling with the finetuned model.

%% file: 4_exp.tex
\section{Experiment Setup}

\paragraph{Dataset}

To evaluate the performance of our proposed method, we conduct experiments on the GeoQuery dataset \cite{zelle1996learning}, where there are 880 queries to a database of U.S. geography. To test compositional generalizability, we adopted the compositional split for SQL released by \citet{finegan2018improving}, where templates created by anonymizing entities are used to split the original dataset, to make sure that all examples sharing a template are assigned to the same set. There are 536/159/182 examples for train/dev/test set, thus this setting can be regarded as the few-shot setting. We also experimented with even fewer training examples (50, 150). 

Besides, we create and release the EcommerceQuery, a new SQL semantic parsing dataset in E-commerce domain. Specifically, we collect natural language utterances from user input search queries to an e-commerce website. To create corresponding SQL queries, we use some self-defined rules with manual audition. We construct compositional splits, where there are unseen SQL query patterns in the dev/test set. Finally, train/dev/test set contains 1,050/353/355 examples respectively. For details, please refer to Appendix~\ref{app:data}. Two examples from EcommerceQuery are shown in Table~\ref{tab:widgets55}.


\paragraph{Baselines and Models} 
We use seq2seq finetuned $\text{BART}$ as our main baseline on both datasets. Without explicit notations, we use $\text{BART}$ large in all of the following experiments. Otherwise, we denote large or base models. On GeoQuery dataset, we use prior state-of-the-art methods as additional baselines. On EcommerceQuery dataset, we use only LSTM seq2seq and $\text{BART}$ as baselines, because \citet{iyer2017learning} requires user feedbacks, and \citet{zheng2020compositional} requires domain specific semantic tags, which are not available in EcommerceQuery.

\paragraph{Evaluation} Following \citet{andreas2019good}, we use exact-match accuracy as the evaluation metric, namely the percentage of examples that are correctly parsed to their SQL queries. 


\section{Experimental Results}

\subsection{Main Results}
\label{main-result}



\begin{table}[tp]
\centering
\small
\setlength\tabcolsep{4pt}
\begin{tabular}{l|c|c}
\toprule
Method                                                       &  GeoQuery & EcoQuery   \\\midrule
\citet{iyer2017learning} $^\dagger$                                         & 40.0  & -         \\
\citet{andreas2019good}  $^\dagger$                               & 49.0    & -        \\
\citet{zheng2020compositional} $^{\dagger\diamond}$          & 69.6   & -     \\
\midrule
\multicolumn{3}{c}{Our Implementation} \\
\hline
$\text{BART}_\text{Base}$                                          &  44.5     & 37.5    \\
$\textsc{SeqZero}_\text{Base}$ &  \textbf{50.0} & \textbf{42.5} \\\cdashline{1-3}
LSTM seq2seq                                    & 39.0     & 9.3      \\
$\text{BART}_\text{Large}$                                         &  72.5     & 37.7    \\
$\text{BART}_\text{Large}$  + SPP & 66.5 & 37.2 \\
$\textsc{SeqZero}_\text{Large}$  &  \textbf{74.7} & \textbf{46.2} \\\bottomrule
\end{tabular}
\caption{\label{tab:widgets1} Results on GeoQuery test set of compositional split, and on EcommerceQuery (EcoQuery) dataset. $^\dagger$: we directly report the metrics in the original papers, while our reproduction achieves similar performance. $^\diamond$: \citet{zheng2020compositional} took an unfair advantage of anonymized variables. }
\end{table}

Table \ref{tab:widgets1} shows our main results on GeoQuery and EcommerceQuery datasets. As shown in Table \ref{tab:widgets1}, on GeoQuery dataset, the finetuned $\text{BART}_{\text{Large}}$ beats all the previous baseline methods. Our approach outperforms all baseline systems by a substantial margin, reaching new SOTA performance. Note that directly combining BART with the semantic parsing via paraphrasing (SSP) framework even decrease the performance of BART, because paraphrased canonical utterances for SQL on GeoQuery is too long and complex to directly generate.
Even comparing with \citet{zheng2020compositional}, {\ours} achieves a much better performance without the usage of anonymized variables \footnote{\citet{zheng2020compositional} could not directly compare with our method, because they use  anonymized variables (i.e. oracle entities), while other models including {\ours} require generating entities instead of using oracle entities. Thus, for fair comparison, their method without variable anonymization would have even worse performance, indicating even larger improvements of our method.}. 
In addition, on EcommerceQuery dataset, our $\textsc{SeqZero}$ further achieves considerable improvements over the baseline methods, reaching SOTA performance. Comparing with BART, the best baseline model, $\textsc{SeqZero}$ gains improvement in exact-match accuracy by $8.5\%$.
In all words, our model is an extremely strong performer and substantially outperforms baseline methods, which demonstrate the efficiency of our method.



\subsection{Ablation Study}

To demonstrate the utility of sequential prompt filling and zero-shot model, we conduct a set of ablation experiments, as shown in Table \ref{tab:widgets15}. In each ablation experiment, we delete one of these two key components of  $\textsc{SeqZero}$, namely ``$-\textsc{Seq}$'' and ``$-\textsc{Zero}$''. 

\begin{table}[!htp]
\centering

\setlength\tabcolsep{4pt}
\begin{tabular}{l|c|c}
\toprule
Method                                                       &  GeoQuery & EcoQuery   \\\midrule
    $\textsc{SeqZero}$ &  \textbf{74.7} & \textbf{46.2} \\
\ \ \    $-\textsc{Seq} $  &  74.2 & 44.5 \\
\ \ \    $-\textsc{Zero} $                                &  71.4 & 37.7  \\\bottomrule
\end{tabular}
\caption{\label{tab:widgets15} Ablation study of $\textsc{SeqZero}$. }
\end{table}
\textbf{{\ours} $-\textsc{Zero}$}
 means that we directly use finetuned few-shot models to fill in sequential prompts without using the zero-shot model.

\textbf{{\ours} $-\textsc{Seq}$} is equivalent to the ensemble of a finetuned BART and a un-finetuned BART for predicting the SQL query directly without sequential prompt filling. 



On both datasets, ``$-\textsc{Seq}$'' decreases the performance of $\textsc{SeqZero}$.
It indicates that designing clause-specific prompt can better mine the pretrained knowledge from the language model.
Meanwhile, zero-shot model ensemble brings our model better out-of-distribution generalization ability. Consequently, when zero-shot model ensemble is ablated, the performance drops a lot (``$-\textsc{Zero}$'' vs ``$\textsc{SeqZero}$'').

\subsection{Analysis of Sequential Prompt Based Models}
\label{seqprompt}
Here, we try to understand how the sequential prompt based model performs on different clauses.
We report the prediction accuracy of $\textsc{SeqZero}$ and ``$- \textsc{Zero}$'' on 5 clauses on the GeoQuery dataset in Table \ref{tab:widgets4}. $\textsc{Seq}_{\text{Gold}}$ means we use finetuned BART to generate clauses given previous gold clauses. We can see that finetuned BART has the worst performance on ``\textsc{From}'' clause because of its poor OOD generalizability.  We can clearly see that $\textsc{SeqZero}$ has better performance than ``$- \textsc{Zero}$'' because of the zero-shot model's strong performance on the ``\textsc{From}'' clause.

\begin{table}[!htp]
\centering
\setlength\tabcolsep{4pt}
\begin{tabular}{@{}l@{}|@{}c@{ }|@{}c@{ }|@{}c@{ }|@{}c@{ }|@{}c@{}}
\toprule

       Method                      & \textsc{From}          & \textsc{Select}        & \textsc{Where}         & \textsc{Group}      & \textsc{Order} \\\midrule
$\textsc{Seq}_{\text{Gold}}$                                       & 84.1          & 87.9          & 92.3          & 99.5          & 99.5               \\\cdashline{1-6}
$\textsc{SeqZero}$  & \textbf{88.5} & \textbf{77.5} & \textbf{74.7} & \textbf{74.7} & \textbf{74.7}   \\
~~~$-\text{Zero}$                     & 84.1          & 74.2          & 71.4          & 71.4          & 71.4               
\\\bottomrule
\end{tabular}
\caption{\label{tab:widgets4} Prediction accuracies on all 5 clauses on GeoQuery dataset.}
\end{table}

Recall that the prediction of the latter clauses depends on the previous ones, the performance of each next clause generally decreases due to error propagation in {\ours}. The same performance of ``\textsc{Where}'', ``\textsc{Group}'' and ``\textsc{Order}'' is because there are very few ``\textsc{Group}'' and ``\textsc{Order}'' clauses on test set. {\ours} achieves much better performance than ``$- \textsc{Zero}$'' on the ``\textsc{From}'' clause and thus significantly reduces the error propagation, leading to better performance on all clauses.

\subsection{Comparison of Zero-shot, Few-shot models, and Their Ensemble}

\begin{table}
\centering
\setlength\tabcolsep{4pt}
\begin{tabular}{c|c}
\toprule

Method                                                               & Exact Match \\\midrule
 \multicolumn{2}{c}{GeoQuery ``\textsc{From}'' Clause} \\\cdashline{1-2}
$\textsc{Few shot}_\text{Base}$  & 58.2          \\
$\textsc{Zero shot}_\text{Base}$        & \textbf{67.0}          \\\cdashline{1-2}
$\textsc{Few shot}_\text{Large}$  & 84.1          \\
$\textsc{Zero shot}_\text{Large}$        & 78.0          \\
$\textsc{Ensemble}_\text{Large}$    & \textbf{88.5} \\\midrule
 \multicolumn{2}{c}{EcommerceQuery ``\textsc{Condition}'' Clause} \\\cdashline{1-2}
$\textsc{Few shot}_\text{Large}$  & 40.0         \\
$\textsc{Ensemble}_\text{Large}$    & \textbf{51.8} 
\\\bottomrule
\end{tabular}
\caption{\label{tab:widgets20} Zero-shot and few-shot $\text{BART}_\text{Base}$ and $\text{BART}_\text{Large}$ models' performance compares with their ensemble on critical clauses. }
\end{table}

According to Section \ref{seqprompt}, our model's major improvement comes from the contribution of zero-shot models and ensemble in critical clauses. We further compare the performance of zero-shot, few-shot and ensemble models in Table \ref{tab:widgets20}. We can see that on GeoQuery  ``\textsc{From}'' Clause, with $\text{BART}_\text{Base}$, zero-shot model itself with constraint decoding is already much better than few-shot model, verifying our intuition that few-shot finetuning could lead model to overfit spurious biases, and achieves poor compositional out-of-distribution (OOD) generalizability. With $\text{BART}_\text{Large}$, zero-shot model's performance is still worse than the few-shot fintuned model, but our ensemble method can effectively leverage the better OOD generalizability of zero-shot model and achieves better performance\footnote{We tried both uncertainty based model selection and model ensemble on ``\textsc{From}'' clause of GeorQuery dataset, and found out that they have similar performance. Thus, we choose model ensemble as our major method, because it leverages all steps' probability to make selection, leading to potentially better performance in other datasets. See Appendix for results of uncertainty based model selection.}. Similarly, on EcommerceQuery ``\textsc{Condition}'' Clause, our ensemble method significantly outperforms the few-shot model.



\subsection{Impact of Prompt Designing}

\begin{table}
\centering
\setlength\tabcolsep{4pt}
\begin{tabular}{c|c|c}
\toprule
Prompt                                   & $\text{Few}$ & $\textsc{Zero}$ \\\midrule
\textit{the answer can be obtained from}  & 81.3                              & 65.9                       \\
\textit{the sentence talks about}       & \textbf{84.1}                     & \textbf{78.0}  

\\\bottomrule
\end{tabular}
\caption{\label{tab:widgets5} Impact of prompt designing for few-shot $\text{Few}$ and zero-shot $\textsc{Zero}$ BART on ``\textsc{From}'' clause of GeoQuery test set.}
\end{table}

Table \ref{tab:widgets5} shows the performance of the few-shot finetuned BART and the zero-shot BART (in constrained decoding setting) with several representative prompts on ``\textsc{From}'' clause of GeoQuery test set. We can see that prompt designing highly affects the the zero-shot model's performance, while it has less impact on few-shot finetuned model. Table \ref{tab:widgets6} shows the performance of the zero-shot BART on ``\textsc{Condition}'' part of EcommerceQuery test set, where different prompts also lead to significantly different performance.
These results reveal the necessity of sequential prompt filling. Without this component, one cannot easily come up with a proper prompt for achieving a better model performance.
In practice, we design 20 prompt sets and select the best one based on the zero-shot model’s performance on the development dataset.

\begin{table}
\centering
\setlength\tabcolsep{4pt}
\begin{tabular}{c|c|c}
\toprule
Prompt  & attribute+relation & relation \\\midrule
\textit{the sentence requires}       & 39.2                & 49.3          \\
\textit{where}              & 21.1                & 51.5          \\
\textit{the condition is :} & \textbf{51.1}       & \textbf{57.3}

\\\bottomrule
\end{tabular}
\caption{\label{tab:widgets6} Impact of prompt designing for zero-shot BART on ``\textsc{Condition}'' clause of EcommerceQuery test set. In attribute+relation setting, we let zero-shot model generate both attributes and relations. In relation setting, we let zero-shot model generate relations only. }
\end{table}

\subsection{Impact of Training Data Size}

\begin{table}
\centering
\setlength\tabcolsep{4pt}
\begin{tabular}{l|c|c|c|c}
\toprule
\# of Samples & 50     & 150  & 536  \\\midrule
$\text{BART}$           & 41.2  & 73.1 & 72.5 \\
$\textsc{SeqZero}$        & \textbf{48.9}  & \textbf{74.2} & \textbf{74.7}\\
~~~~$-\textsc{Zero}$           & 31.3  & 73.1 & 71.4

\\\bottomrule
\end{tabular}
\caption{\label{tab:widgets7} Model accuracy with different numbers of training samples on GeoQuery dataset.}
\end{table}

Table \ref{tab:widgets7} shows the performance of baseline $\text{BART}$ and our $\textsc{SeqZero}$ (as well as ablation of $\textsc{Zero}$), facing different numbers of training data points in the few-shot setting. With 50, 150 training samples, we make sure that each SQL query template occurs only once to maximize the diversity of training data. 
For the full dataset, there are 536 samples with 158 different training templates in total.

Our $\textsc{SeqZero}$ outperforms $\text{BART}$ in all settings (50, 150, 536 training samples), which shows the effectiveness of our method in the few-shot setting. From 50 to 150 training samples, the model see more SQL templates, which help compositional generalization, and lead to the increased performance of all models. From 150 to 536 samples, the performance of \text{BART} and ``$-\textsc{Zero}$'' decrease slightly. That is because there are multiple samples of the same templates in the 536 training samples, and  the models overfit to those training templates.
In contrast, $\textsc{SeqZero}$ avoids such overfitting with the help of zero-shot models and achieves better performance by leveraging more training samples.

Without the aid of zero-shot model, ``$-\textsc{Zero}$'' performs worse than $\textsc{SeqZero}$. When there are only 50 samples, the performance degradation is the most significant. When there are 536 samples, the decrease led by ablation of zero-shot model is larger than that of 150 samples. It is because when there are many cases for each template, ensemble of zero-shot model can alleviate overfitting such templates.

Furthermore, ``$-\textsc{Zero}$'' has similar performance with $\text{BART}$ when there are over 150 training samples. On the other hand, the performance of ``$-\textsc{Zero}$'' is worse than $\text{BART}$ when there are very few training samples (50 samples). We conjecture that this is because $\text{BART}$ shares the model parameter between all sub-clauses, while ``$-\textsc{Zero}$'' finetunes models separately on different sub-clauses. The parameter sharing will further lead to knowledge sharing across sub-clauses and improves the performance. How to leverage the benefit from both parameter sharing and {\ours} could be an interesting future research topic. 

\subsection{Case Study}
\label{case-study}

Table \ref{tab:widgets55} shows BART and \textsc{SeqZero}'s predictions for some cases. For first example, BART gives a wrong prediction, because few-shot training samples introduce too many spurious biases to the finetuned model. In contrast, \textsc{SeqZero} gives correct prediction. Actually, after analyzing the errors made by finetuned $\text{BART}_{\text{Base}}$ model on GeoQuery, among all errors on test set, the common error for around 75\% examples is the table name error in ``\textsc{FROM}'' clause, which is due to spurious biases.


For the second example, $\text{BART}$ predicts ``\textsc{Price <}'' incorrectly even seeing ``\textit{over}'', because EcommerceQuery Dataset is designed to include only ``\textsc{Price <}'' but no ``\textsc{Price >}'' template. Our \textsc{SeqZero} could give the correct prediction because of better OOD generalizability with the help of zero-shot models. 

Even with our \textsc{SeqZero}, there are still many errors. For instance, in the third example, it still struggles with identifying the size in the natural language query and generating the Size filtering condition in \textsc{Where} clause.

\begin{table*}
\small
\centering
\setlength\tabcolsep{4pt}
\begin{tabular}{r|l}
\toprule
                             Cases        & Text \\\midrule
Question      &   \textit{what is the population of utah}   \\
\text{BART} & SELECT city . population FROM city WHERE city . city\_name = "utah" \\
\textsc{SeqZero} & SELECT state . population FROM state WHERE state . state\_name = "utah" \\
Ground Truth & SELECT state . population FROM state WHERE state . state\_name = "utah" \\\midrule


Question      &   \textit{petrol trimmer over 100 dollar}   \\
\text{BART} & SELECT * FROM ASINs WHERE Maching Algorithm(“petrol trimmer”) == True and Price < 100
 \\
\textsc{SeqZero} & SELECT * FROM ASINs WHERE Maching Algorithm(“petrol trimmer”) == True and Price > 100  \\
Ground Truth & SELECT * FROM ASINs WHERE Maching Algorithm(“petrol trimmer”) == True and Price > 100  \\\midrule

Question      &   \textit{mi4 64 gb mobile phone}   \\
\text{BART} & SELECT * FROM ASINs WHERE Maching Algorithm(“mi4 64 gb mobile phone”) ORDER BY date
 \\
\textsc{SeqZero} & SELECT * FROM ASINs WHERE Maching Algorithm(“mi4 64 gb mobile phone”) ORDER BY date  \\
Ground Truth & SELECT * FROM ASINs WHERE Maching Algorithm(“mi4 mobile phone”) and Size = 64 gb   
 \\\bottomrule
\end{tabular}
\caption{\label{tab:widgets55} Case study. The first example is from GeoQuery, and the last two examples are from EcoQuery.}
\end{table*}






%% file: 2_relatedwork.tex
\section{Related Work}

\paragraph{Few/Zero-shot Semantic Parsing} \citet{shin2021constrained, schucher2021power} conducted few-shot semantic parsing by using pretrained LMs to first generate canonical natural language utterances, and then transform them to final formal language through synchronous context-free grammar (SCFG) \cite{jia2016data}. However, dealing with complex structure and lengthy canonical language is still challenging for models in the few-shot setting. Also, canonical languages created through SCFG allows limited space for prompt designing, and canonical language’s form is still too strange for language models to understand. \citet{zhong2020grounded} explored zero-shot semantic parsing via generation-model-based data augmentation. Other ways of bootstrapping a semantic parsing requires rules/grammars to synthesize training examples \cite{xu2020autoqa, wang2015building, yu2020grappa, campagna2019genie, weir2020dbpal, marzoev2020unnatural, campagna2020zero}. \citet{yang2021frustratingly} used language-independent features for zero-shot cross-lingual semantic parsing.

\paragraph{Semantic Parsing via Paraphrasing} \citet{berant2014semantic} started the line of work where semantic parsing is finished through an intermediate paraphrasing step. \citet{wang2015building,marzoev2020unnatural} generated paraphrase candidate values from a grammar of legal canonical utterances, and incrementally filtered the bottom-up or top-down generation by scoring the partial candidates against final formal language. All such work did not exploit the power of pretrained models to generate intermediate paraphrases.

\paragraph{Compositional Generalization in Semantic Parsing} Compositional generalization is an essential problem in semantic parsing because formal languages are internally compositional. Generally, one way to improve compositional generalizability is to incorporate inductive biases directly to models through moduler models \cite{dong2018coarse}, symbolic-neural machines \cite{chen2020compositional}, latent variables/intermediate representations \cite{zheng2020compositional, herzig2020span}, meta-learning \cite{lake2019compositional} etc. Another way is to first do data augmentation and then train a model with augmented data \cite{andreas2019good,zhong2020grounded,yu2020grappa,akyurek2020learning,yang2022subs}. Pretrained models has also been shown useful for compositional semantic parsing \cite{oren2020improving,furrer2020compositional}.  None of prior work used sequential prompts or zero-shot models for compositional generalization. \citet{yang2022tableformer} adopted attention biases to alleviate spurious biases in table semantic parsing.

\paragraph{Prompting for Few/Zero-shot learning} Natural language prompts are widely used in few-shot or zero-shot learning. There are several fashions to use prompts in Autoregressive Language Models \cite{liu2021pre}. One is tuning-free prompting, for example, \citet{petroni2019language,shin2020autoprompt} used a fill-in-the-blank paradigm, while \citet{brown2020language,shin2021constrained} used “few-shot” prompts that included several examples of inputs followed by target outputs, with the actual task input appended at the end. One is Fixed-LM Prompt Tuning, as used by \citet{li2021prefix,schucher2021power,qin2021learning,liu2021gpt}, which requires training less parameters compared with tuning the whole model. Another is Fixed-prompt LM Tuning, which is similar to our setting. We choose to use this way because it is demonstrated better than other methods in many few-shot NLP tasks \cite{gao2020making} when tuning the whole model is not a concern. This is also more efficient at inference time, as it is no longer necessary to select training examples to precede the test input. Note that, \citet{mishra2021reframing} employed prompt decomposition during tuning-free prompting, which is validated in other NLP tasks.

\paragraph{Zero-shot pretrained models for OOD generalization} \citet{wortsman2021robust} showed that, in computer vision tasks, although fine-tuning approaches substantially improve accuracy in-distribution, they reduce out-of-distribution robustness, while zero-shot pretrained models have higher OOD generalizability. Thus, model weight ensemble \cite{wortsman2021robust} and model editing \cite{mitchell2021fast} were leveraged to manipulate zero shot pretrained models, which motivetes us to ensemble zero-shot and few-shot models during the generation process of semantic parsing. We tried weight ensemble proposed by \citet{wortsman2021robust}, but it does not work in our generation setting. The reason is the same as why direct ensemble in prediction space is not working. That’s said, weights in a zero-shot model correspond to the probability over the whole vocabulary while weights in a finetuned model correspond to the probability over constrained vocabulary. Thus, weights in the zero-shot model have little effect on the constrained vocabulary.

%% file: 5_conclusion.tex
\section{Conclusion}
Although prior work leveraged pretrained LMs and canonical language for few-shot semantic parsing, generating lengthy and complex canonical language is still challenging, leading finetuned models to overfitting spurious biases in few-shot training examples and demonstraining poor compositional generalizability. To tackle this, we propose to filling in sequential prompts with LMs and then compose them to obtain final SQL queries. During the process, our proposed zero-shot pretrained model ensemble or uncertainty-based model selection could significantly boost the performance on critical clauses, leading to overall SOTA performance, among BART based models, on GeoQuery and our released EcommerceQuery semantic parsing dataset. In the future, we plan to extend our methods to other pretrained models (e.g. T5) and other compositional semantic parsing datasets.

\section*{Ethical Impact}
{\ours} is a general framework for few-shot semantic parsing on text, such as search queries. {\ours} neither introduces any social/ethical bias to the model nor amplify any bias in the data. When creating EcommerceQuery dataset, we collected data on an E-commerce search platform without knowing customers' identity. No customer/seller specific-data is disclosed. We build our algorithms using public code bases (PyTorch and FairSeq). We do not foresee any direct social consequences or ethical issues.

%% file: appendix.tex



\section{Configuration}
\subsection{Training Details}
During training, we use fairseq \cite{ott2019fairseq} to implement BART model. We use Adam as optimizer with a learning rate 1e-5. We use dropout and attention dropout with 0.1 as dropout rate. Also, we use label smoothing with a rate 0.1. Batch sizes are 1024 tokens. Besides, we employ a weight-decay rate 0.01.
All the parameters are manually tuned based on the dev performance.

We train all models on NVIDIA A100 SXM4 40 GB GPU. We set the max training epoch to be 100 and select the best performed epoch according to dev performance. Training process on each clause or whole sequence could be finished within 3 hours.

\subsection{Inference Details}
During inference, we use greedy search to decode. We also use ensemble of zero-shot and few-shot models during this process. The ensemble weight $\gamma_i$ in Eq.~\eqref{eq:ensemble} is chosen from [0, 1] and tuned by grid search according to performance on dev set. 

\section{EcommerceQuery Dataset}
\label{app:data}

When we create the EcommerceQuery dataset, we first we collect natural language utterances from user input search queries to an e-commerce website. To create corresponding SQL queries, we use regular expressions to create ``\textsc{Size}'' filtering conditions, and use some rules to create ``\textsc{Price}'' filtering conditions, ``\textsc{Delivery}'' attributes and ``\textsc{Subscribe}'' attributes in ``\textsc{Where}'' clauses. Finally, we manually audit each pair of data to ensure the quality. 


To construct compositional splits, we make sure that there is no ``\textsc{Price>}'', ``\textsc{Size=}'', and ``\textsc{Subscribe=}'' SQL templates in training set but the majority of SQL queries on dev and test set contains such templates. Ideally, a model with good compositional generalizability could generalize from ``\textsc{Price<}'' and ``\textsc{Size>}'' to ``\textsc{Price>}'', generalize from ``\textsc{Price=}'' and ``\textsc{Size>}'' to ``\textsc{Size=}'', and generalize from ``\textsc{Delivery=}'' to ``\textsc{Subscribe=}''. 


\section{Problem Decomposition on GeoQuery and EcommerceQuery}
\label{app:compose}

In this section we introduce the problem decomposition for GeoQuery and EcommerceQuery in details. We answer the following two questions: 1. what are the sub-clauses in the sub-problems? 2. how to compose the final formal language from the sub-clauses. 

\subsection{GeoQuery}

On GeoQuery, there are totally 5 sub-clauses, namely \textsc{From}, \textsc{Select}, \textsc{Where}, \textsc{Group-By}, \textsc{Order-By} clauses. we first generate \textsc{From} from clause with the prompt ``\textit{the sentence talks about}''. Then we generate \textsc{Select} clause with the prompt ``\textit{the sentence talks about}'', generate ``\textit{Where} clause with the prompt \textsc{the sentence requires}'', generate \textsc{Group-By} clause with the prompt \textsc{the sentence requires to group by}, and generate \textsc{Order-By} clause with the prompt ``\textit{the sentence requires the result to be ordered by}'' Note that prior generated clauses are used as additional prefix to generate current clauses. The filled value for each clause could be ``\textit{None}''. When the filled value is ``\textit{None}'', which means there is no such clause in the final SQL query. Finally, we compose all clauses (if the filled value is not ``\textit{None}'') sequentially to obtain the final SQL query.

\subsection{EcommerceQuery}
On EcommerceQuery, there are totally 2 sub-clauses, namely \textsc{Matching}, and \textsc{Condition} clauses. Because thes two clauses are less dependent, we generate each clause separately and then compose the generated values of each clause. When generating \textsc{Matching} clause, we use the prompt ``\textit{matching algorithm (}''. When generating \textsc{Condition} clause, we use the prompt ``the condition is :''.

\section{Uncertainty based Model Selection}
\begin{table}
\centering
\setlength\tabcolsep{4pt}
\begin{tabular}{c|c}
\toprule

Method                                                               & Exact Match \\\midrule
$\textsc{Few shot}_\text{Large}$   & 84.1          \\
$\textsc{Zero shot}_\text{Large}$        & 78.0          \\\cdashline{1-2}
$\textsc{MoC Selection}_\text{Large}$    & \textbf{88.5} \\
$\textsc{RoC Selection}_\text{Large}$     & \textbf{88.5} \\
$\textsc{Ensemble}_\text{Large}$     & \textbf{88.5} 
\\\bottomrule
\end{tabular}
\caption{\label{tab:widgets22} Ensemble of zero-shot and few-shot models compares with uncertainly based selection of zero-shot and few-shot models on GeoQuery ``\textsc{From}'' Clause. }
\end{table}
As an alternative to model ensemble, we can also decide whether to use the predicted sequence of the zero-shot model or the fine-tuned model based on zero-shot model's uncertainty score over the generated sequence. Specifically, during greedy search, we compute an uncertainty metric with the rescaled zero-shot model prediction  $p^{\star T}$, where $T$ is the first decoding step after the pre-designed prompt \footnote{The reason why we choose $T$th step is that we do not want to consider the probability of $\textsc{[EOS]}$ token into uncertainty, because for most table name tokens, there is little probability that the $\textsc{[EOS]}$ token occurs after them in zero-shot models.}.  The uncertainty metric could be Margin of Confidence ($\textsc{MoC}$)  or Ratio of Confidence ($\textsc{RoC}$) . Formally, assume the largest value in vector $p^{\star T}$ is $p^{\star T}_1$, and the second largest value in vector $p^{\star T}$ is $p^{\star T}_2$, we compute these two uncertainty metrics as:

\begin{equation}
\label{eqtatt}
\begin{aligned}
\textsc{MoC}&=1-(p^{\star T}_1 - p^{\star T}_2) \\
\textsc{RoC}&=p^{\star T}_2 / p^{\star T}_1
\end{aligned}
\end{equation}

The results are shown in Table \ref{tab:widgets22}.